\documentclass[a4paper,fleqn]{cas-dc}

\usepackage[authoryear,longnamesfirst]{natbib}

\usepackage{tikz}
\usetikzlibrary{positioning,arrows.meta,calc,fit,backgrounds}
\usepackage{pgfplots}
\pgfplotsset{compat=1.17}

\usepackage{lmodern}
\ifdefined\pdfgentounicode
  \input{glyphtounicode}
\fi

\begin{document}
\let\WriteBookmarks\relax
\renewcommand{\topfraction}{.95}
\renewcommand{\bottomfraction}{.90}
\renewcommand{\textfraction}{.05}
\renewcommand{\floatpagefraction}{.70}
\renewcommand{\dbltopfraction}{.95}
\renewcommand{\dblfloatpagefraction}{.70}
\setcounter{topnumber}{5}
\setcounter{bottomnumber}{5}
\setcounter{totalnumber}{10}
\setcounter{dbltopnumber}{3}

\shorttitle{LDAC-Net for Low-Cost MOX Gas-Sensor Recognition}

\providecommand{\papertitle}{LDAC-Net: A Learnable Multi-Lag Differencing Attention-Convolution Network for Drift-Robust Recognition with Low-Cost MOX Gas Sensors}
\title[mode=title]{\papertitle}

\ifdefined\ANONYMIZED
\shortauthors{Anonymous}
\hypersetup{pdfauthor={Anonymous}}
\else
\shortauthors{X. Zhang et~al.}
\author[1]{Xin Zhang}

\author[1]{Liangxiu Han}
\cormark[1]
\ead{l.han@mmu.ac.uk}

\author[1]{Yue Shi}

\author[1]{Tam Sobeih}

\affiliation[1]{organization={Department of Computing and Mathematics, Manchester Metropolitan University},
            city={Manchester},
            postcode={M1 5GD},
            country={U.K.}}

\cortext[1]{Corresponding author: Liangxiu Han}
\fi

\begin{abstract}
Portable electronic-nose systems based on low-cost metal-oxide (MOX) gas sensors offer a practical solution for gas and odour recognition, but their signals are affected by slow chemical transients, drifting sensor offsets, scale variation, and cross-channel correlations. Existing pipelines commonly use fixed first-order temporal differencing (FOTD), which requires a manually selected lag and may discard useful response information. We propose \textbf{LDAC-Net}, an end-to-end learnable multi-lag differencing attention-convolution network that operates directly on multi-channel MOX signals. Its learnable differential feature enhancement front-end combines window-conditioned statistical affine normalisation, which compensates for window-specific offset and scale variation, with learnable multi-lag differencing, which weights and combines temporal differences across multiple lags. A compact attention-convolution backbone subsequently models local transients and longer-range temporal dependencies. On the 50-class SmellNet-Base task, LDAC-Net achieves $68.2\,\%$ top-1 accuracy, exceeding the best FOTD-preprocessed comparison model by approximately $14$ percentage points and the raw-input Transformer by more than $30$ points. Ablation studies confirm the contributions of both proposed components. The representation also transfers to SmellNet-Mixtures, improving accuracy from $45.4\,\%$ to $50.5\,\%$, and generalises to the $62$-channel eNose-Drift benchmark under strong long-term drift, achieving $70.6\,\%$ top-1 accuracy and $69.6\,\%$ macro-F1. These results outperform the best comparison model with dataset-retuned FOTD preprocessing by $8.0$ and $3.0$ points, respectively, demonstrating that learnable, sensor-aware preprocessing is more effective than fixed handcrafted differencing for low-cost MOX gas-sensor recognition.
\end{abstract}



\maketitle

\section{Introduction}
\label{sec:intro}

Portable electronic-nose systems built around low-cost metal-oxide (MOX) gas sensors are inexpensive, compact, and well suited to deployment outside the laboratory, and are increasingly used in food quality assessment, indoor air monitoring, odour recognition, and consumer-grade health screening~\citep{poeta2025moxreview,rabehi2024enosereview}. Automatic substance recognition from these sensors is a common but difficult problem, the signals drift across sessions, span multiple temporal scales, and are usually cleaned by fixed handcrafted preprocessing that sits outside the learned model. For MOX in particular, reliable recognition remains hard because the signals are dominated by slow chemical transients, session-dependent baseline drift, scale variation, and strong cross-channel correlations, so the difficulty lies in the slow, drifting, multi-scale dynamics of the signal.

Early electronic-nose studies extracted handcrafted descriptors of the response curve, such as steady-state level and transient shape, for shallow classifiers like linear discriminant analysis or small neural networks~\citep{persaud1982analysis,yan2015enose}. Other studies instead clean the raw signal with fixed preprocessing before classification. First-order temporal differencing (FOTD), used in the recent SmellNet pipeline~\citep{feng2026smellnet}, replaces each sample by an earlier-lagged difference such as $x[t]-x[t-25]$ to suppress slow drift and expose response dynamics. This helps, but it fixes the lag as a hand-picked hyperparameter, applies the same operation to every substance, channel, and session, discards the absolute response level that itself carries class information, and stays outside the model. Other strategies, such as baseline correction or a separate drift-calibration stage~\citep{vergara2012drift,zhang2022tdacnn}, share these limits. Drift compensation and dynamics extraction are thus decoupled from recognition rather than learned jointly with it.

Deep learning offers an alternative to such fixed pipelines and has been widely adopted for multivariate time-series recognition~\citep{ruiz2021bakeoff,foumani2024convtran}, through convolutional networks~\citep{bai2018tcn}, recurrent networks~\citep{hochreiter1997lstm}, and Transformers~\citep{vaswani2017attention,zerveas2021transformer,liu2021gated}, and similar models have been applied to gas-sensor recognition~\citep{peng2018gascnn}. Such models require large training sets, and their use in this domain has been enabled by the release of public gas-sensor corpora~\citep{vergara2012drift,fonollosa2015reservoir,huerta2016homeactivity}, most recently the large-scale SmellNet benchmark~\citep{feng2026smellnet}, which provides $828\,000$ timesteps over $50$ substances on a six-channel MOX array. Even with such data, however, existing models are applied \emph{on top of} FOTD-preprocessed signals and improve only the classifier backbone, while methods that target input non-stationarity, such as reversible instance normalisation~\citep{kim2022revin} and the non-stationary Transformer~\citep{liu2022nonstationary}, normalise statistics away without learning the multi-scale temporal differencing that exposes a slow MOX response. Its reference model, ScentFormer, reaches only $56.1\,\%$ Top-1 on the 50-way base task, and no end-to-end approach yet learns drift compensation and dynamics extraction jointly with recognition.

We propose \textbf{LDAC-Net}, an end-to-end network that classifies raw multi-channel windows directly, with no external preprocessing. Its core is a \textbf{Learnable Differential Feature Enhancement (LDFE)} front-end that performs window-adaptive normalisation and multi-lag temporal differencing inside the network, turning the two effects FOTD relies on, drift removal and dynamics exposure, into learnable operations on the raw window. Unlike normalisation-based approaches to non-stationarity such as AdaIN~\citep{huang2017adain}, DAIN~\citep{passalis2020dain}, RevIN~\citep{kim2022revin}, and the non-stationary Transformer~\citep{liu2022nonstationary}, which only re-scale input statistics, LDFE also learns how to difference the signal across multiple lags to expose its dynamics, and it applies broadly to drift-affected, multi-scale sensor time series rather than to MOX alone. The compensated signal is read by a compact \textbf{attention-convolution (AC)} backbone and an attention-pooling head; we adopt this hybrid rather than a plain Transformer because such sensor windows require both local transient modelling and longer-range temporal context, while the available training data remain limited~\citep{gulati2020conformer}. We further identify the analysis window length as a strong but previously unexploited factor, since a slow MOX transient needs sufficient temporal context before in-model differencing becomes informative. This paper makes three main contributions.

\begin{itemize}
  \item We propose \textbf{LDAC-Net}, an end-to-end network for substance recognition on low-cost MOX gas-sensor arrays. By learning drift compensation, dynamics extraction, and classification jointly, LDAC-Net avoids traditional manual preprocessing or recalibration.
  \item We propose \textbf{LDFE}, the learnable front-end at its core, built from two parameter-light, per-window stages: \textbf{WSAN}, whose zero-initialised MLP predicts per-channel affine corrections from each window's own mean and standard deviation, and \textbf{LMLD}, which forms temporal differences over a bank of lags and learns their per-channel weighting, cross-channel mixing, and gating.
  \item We identify the analysis window length as a strong and previously unexploited factor: a longer window gives the in-model differences enough temporal context to expose the response dynamics rather than noise.
\end{itemize}

\section{Related Work}
\label{sec:related}


\subsection{Low-Cost MOX Gas-Sensor Recognition}

Low-cost metal-oxide (MOX) gas sensors are common in electronic-nose systems because they are cheap, compact, and portable. Early studies used small, task-specific corpora, such as coffee-defect detection~\citep{rodriguez2010coffee}, beef-freshness monitoring~\citep{wijaya2018beef}, and the Gas Sensor Array Drift dataset~\citep{vergara2012drift}, typically extracting handcrafted features from the response curve for a shallow classifier~\citep{persaud1982analysis,yan2015enose}; such features are sensitive to drift and acquisition variation. More recent work swaps in deep networks~\citep{peng2018gascnn} but still relies on conventional drift-removal preprocessing upstream. The SmellNet benchmark~\citep{feng2026smellnet} scales the problem up (six channels, 50 base substances, 43 mixtures, 828\,000 timesteps over 68 hours), yet its Transformer reference (ScentFormer) reaches only $56.1\,\%$ Top-1 on the 50-way task. Recognition stays hard because MOX responses are slow, cross-sensitive, and drift-affected, which calls for models that handle these signal properties rather than just stronger classifiers.

\subsection{Drift Compensation and Temporal Differencing}

The core difficulty is that the signal mixes class-relevant response information with nuisance variation from baseline drift, scale changes, and session-dependent offsets, so e-nose pipelines preprocess before classification. Temporal differencing addresses this directly: subtracting an earlier reading suppresses slow drift and exposes the transient dynamics, much like the delta (dynamic) features used in speech recognition~\citep{furui1986delta}. SmellNet uses first-order temporal differencing (FOTD), a fixed operator with a hand-picked lag that is applied to every substance, channel, and session and that discards the absolute level, which itself carries class information. Other drift-compensation methods, from baseline correction and channel ratio/difference features~\citep{vergara2012drift} to domain-adaptation networks~\citep{zhang2022tdacnn}, likewise rest on handcrafted assumptions or treat drift as a separate calibration module, and still presuppose a fixed differencing front.

\subsection{Deep Learning for Multivariate Time-Series Recognition}

Deep networks now dominate multivariate time-series classification~\citep{ismailfawaz2019review}. Convolutional models capture local patterns efficiently, from dilated causal stacks (TCN)~\citep{bai2018tcn} and Inception-style ensembles~\citep{ismailfawaz2020inceptiontime} to random convolutional kernels (ROCKET)~\citep{dempster2020rocket}; recurrent models such as LSTM~\citep{hochreiter1997lstm} are compact, while Transformers~\citep{vaswani2017attention} capture long-range structure but are data-hungry and overfit on small sensor corpora, which has prompted efficient variants~\citep{zhou2021informer}. Hybrid convolution-attention backbones~\citep{gulati2020conformer} pair local and long-range modelling, which suits MOX windows, and self-supervised objectives help when labels are scarce~\citep{oord2018cpc}. A parallel line removes input non-stationarity by normalisation, from Instance Normalisation~\citep{ulyanov2016instancenorm} and its input-conditioned variants AdaIN~\citep{huang2017adain} and RevIN~\citep{kim2022revin} to DAIN~\citep{passalis2020dain}, which predicts per-feature shift and scale parameters from each window's summary statistics, and the Non-stationary Transformer~\citep{liu2022nonstationary}. These models normalise statistics but do not learn the multi-scale temporal differencing that exposes a slow MOX response.

Two challenges therefore remain for low-cost MOX recognition: (1) drift compensation and dynamics extraction are handled by fixed handcrafted preprocessing outside the model rather than learned jointly with recognition; and (2) methods that target input non-stationarity normalise statistics away without learning the multi-scale differencing that exposes slow MOX dynamics. LDAC-Net addresses both with a learnable front-end whose novelty is the combination, inside an end-to-end model, of window-conditioned affine correction and learnable multi-lag differencing. This sets it apart from instance-normalisation methods (IN~\citep{ulyanov2016instancenorm}, AdaIN~\citep{huang2017adain}, DAIN~\citep{passalis2020dain}, RevIN~\citep{kim2022revin}) and the Non-stationary Transformer~\citep{liu2022nonstationary}, which only re-scale statistics; from fixed delta features~\citep{furui1986delta} and multi-scale convolutional difference operators~\citep{bai2018tcn,ismailfawaz2020inceptiontime}, which apply pre-defined kernels; and from electronic-nose drift compensation~\citep{vergara2012drift,zhang2022tdacnn} and learnable signal front-ends~\citep{ravanelli2018sincnet,zeghidour2021leaf}, which keep calibration separate from the classifier.

\section{Method}
\label{sec:method}

\subsection{Overview}
\label{sec:method:overview}

In this work, we consider the low-cost MOX gas-sensor recognition as a multivariate time-series classification problem. A session-level recording $\mathbf{X}\in\mathbb{R}^{L\times C}$ of session-specific length $L$ and $C$ channels is segmented along the time axis into fixed-length windows $\mathbf{x}\in\mathbb{R}^{T\times C}$, and the goal is to predict the substance label $y$ directly from a raw window, without external drift correction or handcrafted temporal differencing.

\begin{figure*}[pos=!tbp]
  \centering
  \includegraphics[width=\textwidth]{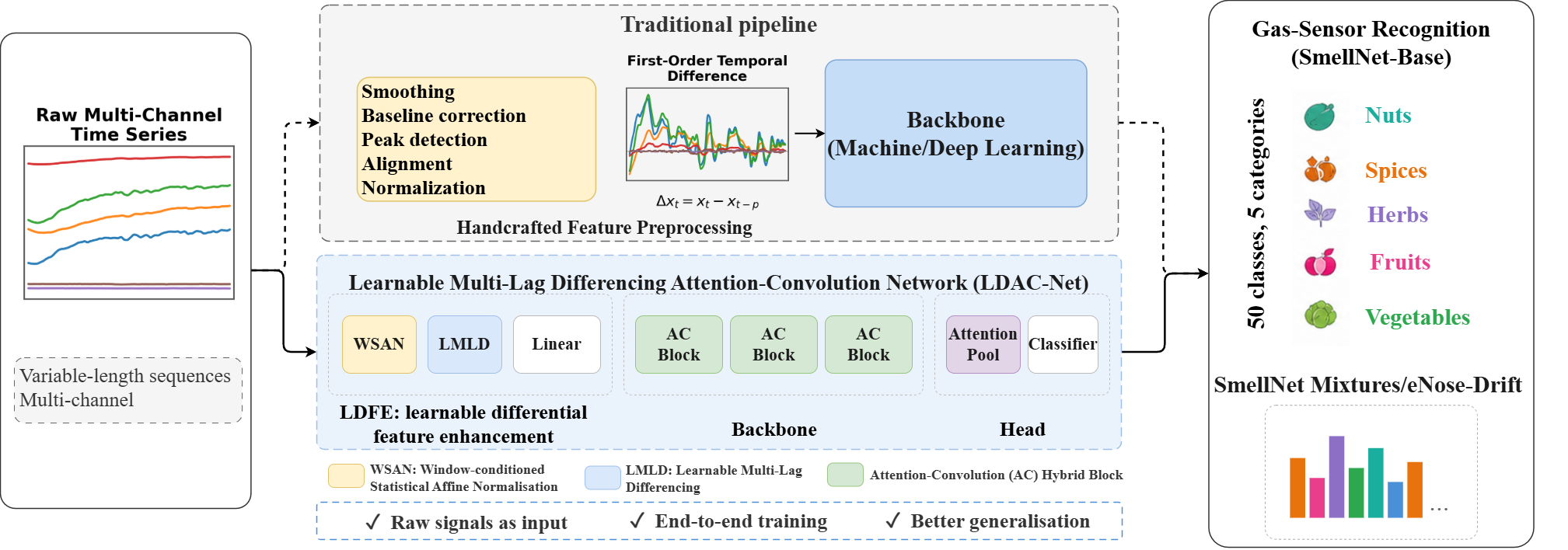}
  \caption{The LDAC-Net architecture. A raw window $\mathbf{x}$ first passes through the LDFE front module (WSAN normalisation and LMLD differencing), is then processed by three AC blocks, and is finally read out by the Attention-Pooling head into the 50-way logits.}
  \label{fig:arch}
\end{figure*}

We propose LDAC-Net, which learns sensor drift compensation and temporal dynamics extraction inside the network end to end. As shown in Figure~\ref{fig:arch}, it is built from three modules. 1)~The Learnable Differential Feature Enhancement (LDFE) front-end compensates the raw signal for per-session drift and exposes its temporal dynamics through two stages, Window-conditioned Statistical Affine Normalisation (WSAN) and Learnable Multi-Lag Differencing (LMLD). 2)~The Attention-Convolution (AC) blocks extract features from the embedded sequence, each combining a depthwise convolution for local structure with multi-head self-attention for long-range context. 3)~The Attention-Pooling head reads the sequence out: a single learnable query attends over the tokens to pool them into a fixed-size embedding, which a linear classifier maps to the class logits. The rationale for each module addresses a generic difficulty of learning from multi-channel time-series signals:
\begin{itemize}
  \item \textbf{Compensate drift and reveal the temporal dynamics of the data.} Real-world sensor recordings drift in baseline and scale across sessions, so the same input statistics rarely recur at test time; at the same time, for such slowly varying signals the discriminative cue often lies in how the signal changes over time rather than in its absolute level. The LDFE front module handles both inside the network. WSAN folds the normalisation in and conditions it on each window's own statistics, so the model adapts to per-window shifts on the fly rather than relying on fixed external preprocessing or a global rule. LMLD then makes the dynamics explicit through learnable multi-lag differencing instead of a single hand-picked lag, keeping the parameter count low by learning only how to weight and combine closed-form differences.
  \item \textbf{Capture local and long-range dependencies jointly.} A signal window carries both short local patterns and dependencies spread across the whole window, so the AC block pairs a depthwise convolution for the local shape with self-attention for long-range context, deliberately scaled down to remain trainable on a small dataset without overfitting.
  \item \textbf{Weight timesteps by their importance.} Timesteps in a window are not equally informative, so the Attention-Pooling head replaces a uniform mean pool with a single-query attention pool that concentrates the read-out weight on the most discriminative parts of the sequence.
\end{itemize}

\subsection{LDFE: Learnable Differential Feature Enhancement}
\label{sec:method:ldfe}

The LDFE module provides the backbone with a representation that is drift compensated and dynamics aware: it removes the slow per-session baseline drift that would swamp the class signal, and surfaces the response dynamics. There are two learnable stages, a window-conditioned affine normalisation (WSAN) and a multi-lag differencing stage (LMLD) respectively, followed by a linear projection that lifts the result to the embedding ($d=128$) the AC blocks consume.

\subsubsection{Window-conditioned Statistical Affine Normalisation (WSAN)}
\label{sec:method:wsan}

\begin{figure}[pos=!tbp]
  \centering
  \includegraphics[width=\columnwidth]{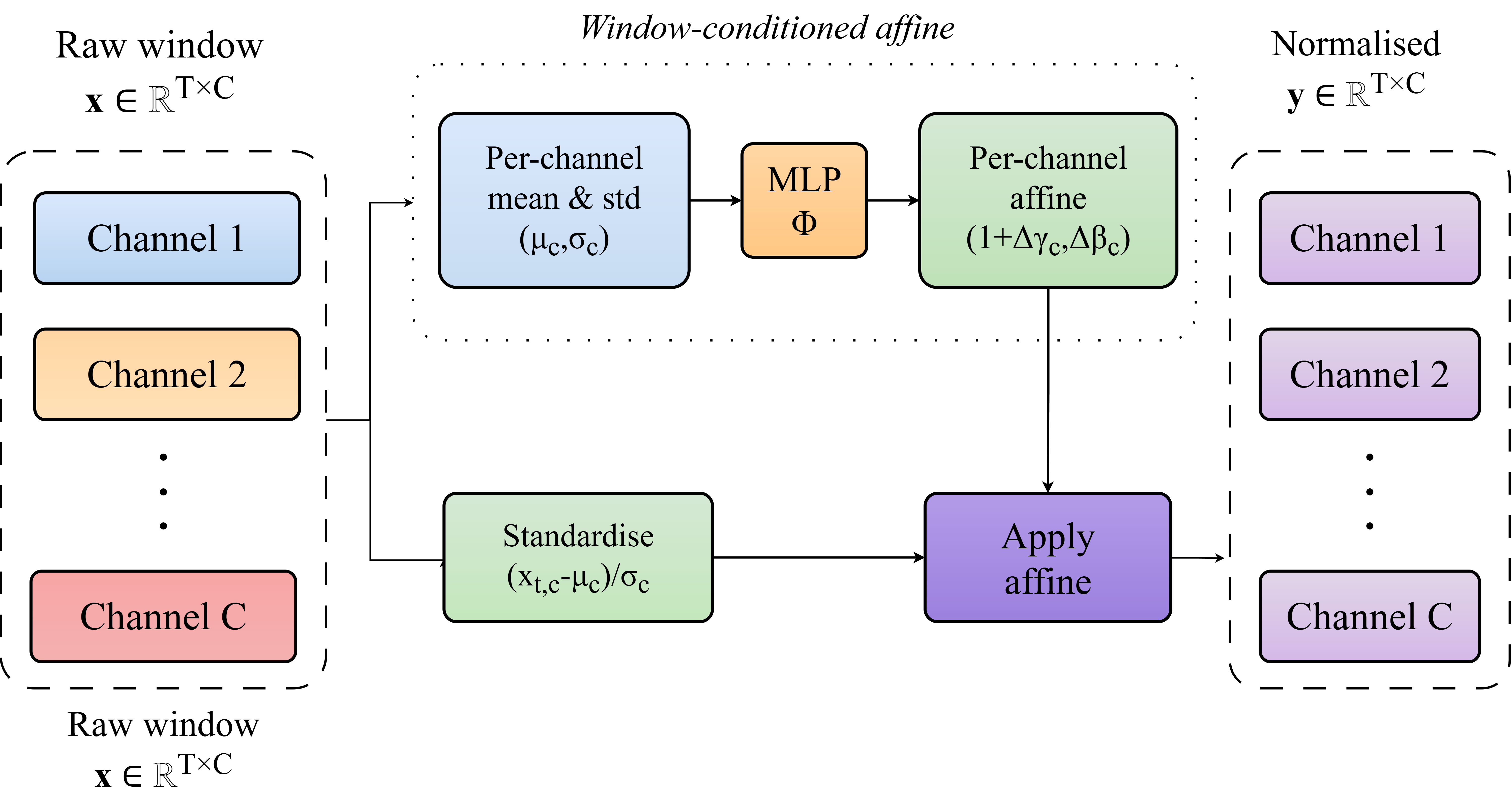}
  \caption{The WSAN module. For each multi-channel window, the per-channel mean and standard deviation feed a small MLP that emits per-channel affine offsets $(1{+}\Delta\boldsymbol{\gamma},\Delta\boldsymbol{\beta})$, which are applied to the standardised window to produce the normalised output.}
  \label{fig:wsan}
\end{figure}

The role of WSAN is to remove per-session baseline drift by normalising each window with an affine transform conditioned on the window's own statistics. Figure~\ref{fig:wsan} shows the WSAN module. For a window $\mathbf{x}\in\mathbb{R}^{T\times C}$, WSAN first computes, for each channel, the mean and standard deviation over time:
\begin{equation}
  \mu_c \;=\; \frac{1}{T}\sum_{t=1}^{T} x_{t,c},
  \qquad
  \sigma_c \;=\; \sqrt{\frac{1}{T-1}\sum_{t=1}^{T} (x_{t,c}-\mu_c)^2},
  \label{eq:wsan-stats}
\end{equation}
with $\sigma_c$ clamped below at $10^{-5}$ for numerical stability. The concatenated statistics $[\boldsymbol{\mu}, \boldsymbol{\sigma}]$ are passed through a two-layer MLP $\Phi$ with a GELU activation and a zero-initialised final layer:
\begin{equation}
  (\Delta\boldsymbol{\gamma},\,\Delta\boldsymbol{\beta})
  \;=\;
  \Phi\bigl([\boldsymbol{\mu},\boldsymbol{\sigma}]\bigr).
  \label{eq:wsan-mlp}
\end{equation}
The WSAN output is the per-window affine instance normalisation
\begin{equation}
  \tilde{x}_{t,c}
  \;=\;
  \bigl(1 + \Delta\gamma_c\bigr)\,
  \frac{x_{t,c} - \mu_c}{\sigma_c}
  \;+\;
  \Delta\beta_c.
  \label{eq:wsan-out}
\end{equation}

WSAN works in three parts: it standardises each channel by its window statistics, predicts a residual per-channel affine from those same statistics, and applies that affine to the standardised signal. Because the final layer of $\Phi$ is zero-initialised ($\Delta\boldsymbol{\gamma} = \Delta\boldsymbol{\beta} = \mathbf{0}$), WSAN starts as exact instance normalisation and gradually learns its per-window correction from data, and because $\Phi$ maps the $2C$ window statistics to $2C$ affine parameters, it adds very few parameters. The affine path is what separates WSAN from plain instance normalisation, which discards $[\boldsymbol{\mu},\boldsymbol{\sigma}]$ together with the drift even though the absolute response level of a MOX array itself carries class information: $\Phi$ re-injects the discriminative part of these statistics through $(\Delta\boldsymbol{\gamma},\Delta\boldsymbol{\beta})$, so WSAN acts as a \emph{leaky} instance normalisation that removes the drift while keeping a learned route for the absolute level.

\subsubsection{Learnable Multi-Lag Differencing (LMLD)}
\label{sec:method:lmld}

In gas-sensor recognition tasks, a key feature to capture is the response dynamics, that is, how fast each channel rises and falls, rather than the absolute signal level~\citep{vergara2012drift}. We therefore design LMLD to capture these dynamics from the raw input by computing temporal differences at several lags.

\begin{figure}[pos=!tbp]
  \centering
  \includegraphics[width=\columnwidth]{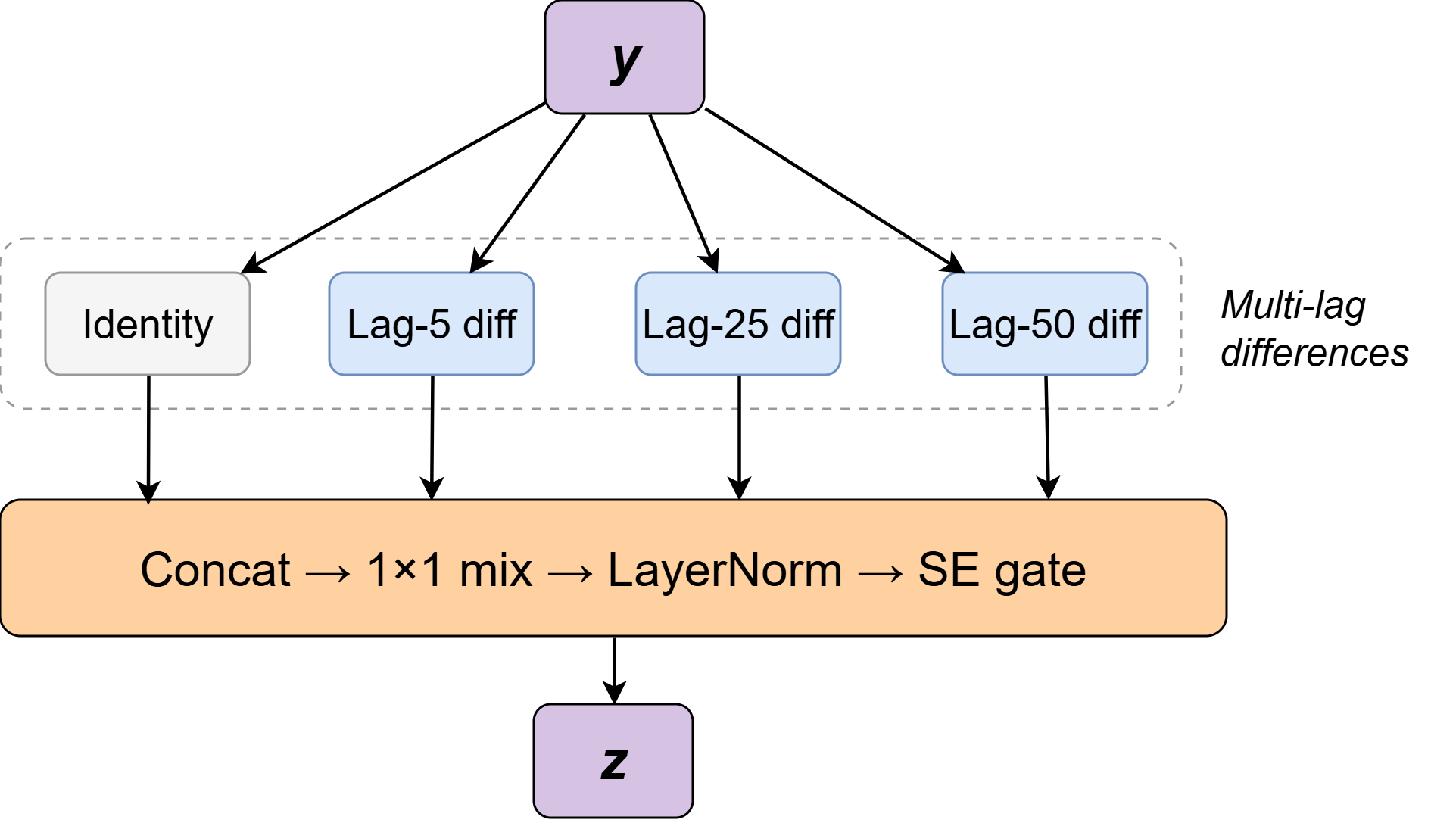}
  \caption{The LMLD module. The WSAN-normalised window is differenced at a bank of lags $P=\{5,10,25,50\}$ (only $5,25,50$ drawn); each difference is scaled by a learnable per-channel weight, concatenated with the identity stream, mixed by a $1{\times}1$ convolution, layer-normalised, and gated by a squeeze-excitation block.}
  \label{fig:lmld}
\end{figure}

Let $\tilde{\mathbf{x}}\in\mathbb{R}^{T\times C}$ be the WSAN output and $P=\{5,10,25,50\}$ a fixed bank of lags. For each lag $p\in P$ LMLD forms the in-window first difference and rescales it by a learnable per-channel weight $\mathbf{w}^{(p)}\in\mathbb{R}^{C}$, initialised to $\mathbf{1}$ so that training starts from plain differences,
\begin{equation}
  \mathbf{d}^{(p)}_{t}
  \;=\;
  \mathbf{w}^{(p)} \odot \bigl(\tilde{\mathbf{x}}_{t} - \tilde{\mathbf{x}}_{t-p}\bigr),
  \qquad \mathbf{d}^{(p)}_{t}=\mathbf{0}\ \text{for}\ t\le p,
  \label{eq:lmld-diff}
\end{equation}
where $\odot$ is the per-channel product. The identity stream and all difference streams are concatenated along the channel axis, mixed by a pointwise ($1{\times}1$) convolution to $d'=64$ channels, layer-normalised, and passed through a squeeze-excitation channel gate $\mathrm{SE}(\cdot)$~\citep{hu2018squeeze} with reduction ratio $8$:
\begin{equation}
  \begin{aligned}
    \mathbf{h} &= \bigl[\,\tilde{\mathbf{x}},\,\mathbf{d}^{(5)},\,\mathbf{d}^{(10)},\,\mathbf{d}^{(25)},\,\mathbf{d}^{(50)}\,\bigr], \\
    \mathbf{z} &= \mathrm{SE}\bigl(\mathrm{LN}(\mathrm{Conv}_{1\times1}(\mathbf{h}))\bigr).
  \end{aligned}
  \label{eq:lmld-mix}
\end{equation}
The output $\mathbf{z}\in\mathbb{R}^{T\times d'}$ replaces the raw window as the input to the linear projection (Figure~\ref{fig:lmld}).

By fixing the temporal differences to a closed form and learning only their weighting, cross-channel mixing, and gating, LMLD generalises a fixed first-order difference into a learnable, multi-lag operator: the per-channel weight of each lag, the cross-channel mixing (the $1{\times}1$ convolution, which subsumes the classic channel-difference feature as a special case), and the channel gating are all learned rather than hand-set. It therefore extracts the response dynamics at several time scales at once, a short lag captures fast transients while a long lag captures the slow rise-and-settle behaviour, and the squeeze-excitation gate emphasises the lags and channels that carry the most useful dynamic information.

\subsubsection{Channel Projection Stage}

A pointwise linear projection lifts the gated, mixed representation $\mathbf{z}\in\mathbb{R}^{T\times d'}$ to the $d$-dimensional token embedding that the AC blocks operate on. This projection closes the LDFE module: a raw window enters and a normalised, dynamics-enhanced $T\times d$ token sequence leaves, with no external preprocessing stage anywhere in the pipeline.

\subsection{Attention-Convolution (AC) Hybrid Block}
\label{sec:method:block}

\begin{figure}[pos=!tbp]
  \centering
  \includegraphics[width=\columnwidth]{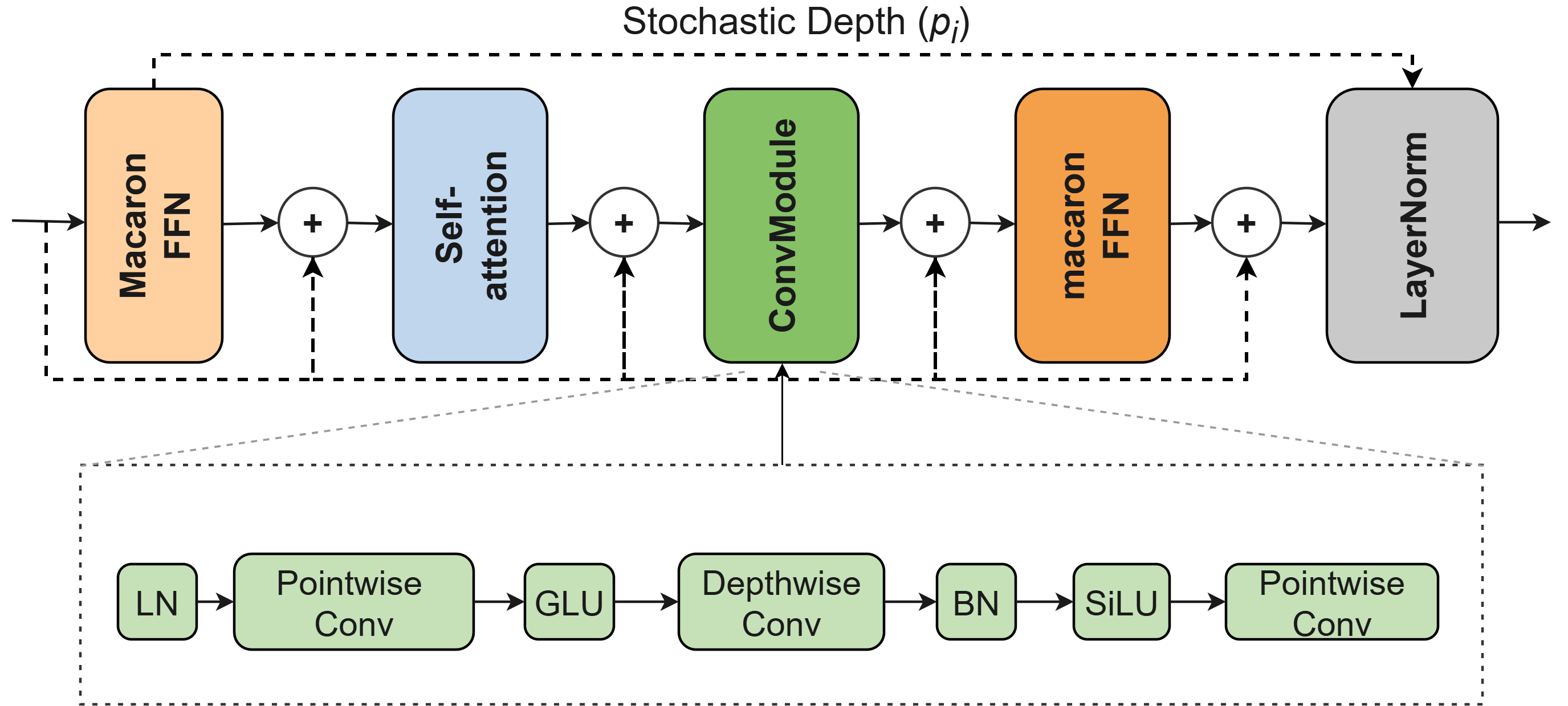}
  \caption{The AC block: two half-step macaron feed-forward layers around a multi-head self-attention and a ConvModule, with per-step residual connections (dashed) and a final LayerNorm. A whole-block identity bypass realises stochastic depth, and the inset expands the ConvModule.}
  \label{fig:cachan-block}
\end{figure}

The AC block is the backbone unit that processes the LDFE output. It is an attention-convolution hybrid: each AC block applies a half-step macaron feed-forward layer, multi-head self-attention, a ConvModule, and a second half-step macaron feed-forward layer, with a final LayerNorm (Figure~\ref{fig:cachan-block}). This structure adopts the Conformer block~\citep{gulati2020conformer}, scaled down for the short multivariate windows and limited training data of gas-sensor recognition. The macaron feed-forward pattern places two residual pre-norm feed-forward blocks (LayerNorm, linear expansion, SiLU, linear projection), each scaled by $0.5$, before and after the attention; sandwiching the attention between two half-step feed-forward layers refines the token representation on both sides of the global mixing. The self-attention uses multi-head scaled dot-product attention to relate distant timesteps and weight which parts of the window are informative. The ConvModule applies a LayerNorm, a pointwise convolution that expands the channels, a GLU gate, a depthwise convolution, a BatchNorm, a SiLU activation, and a final pointwise convolution back to the model width; its local kernel is short enough to detect fast transients yet long enough to integrate across them. The backbone uses no explicit positional encoding: temporal order is conveyed by the depthwise convolutions.

To prevent the deep model from overfitting on the small training set, we apply stochastic depth~\citep{huang2016stochdepth} per block, with a survival probability that decreases linearly with depth:
\[
  p_i \;=\; 1 - p_{\max}\,\frac{i}{N},
  \qquad i = 1,\dots,N,
\]
where $N$ is the number of blocks and $p_{\max}$ the maximum drop rate, so block $i$ is skipped (replaced by the identity) with probability $p_{\max}\,i/N$ during training. Deeper blocks are thus more likely to be dropped during training, which acts as an implicit ensemble over sub-networks of varying depth and reduces the gap between training and test accuracy in the small-data regime. At test time every block fires deterministically.

\subsection{Attention-Pooling Head}
\label{sec:method:head}

The head turns the backbone's $T\times d$ token sequence into class logits in two steps: a single-query attention pool that collapses the time axis, followed by a linear classifier.

The single-query multi-head attention pool uses a learnable query token $\mathbf{q}\in\mathbb{R}^{d}$ that attends over the $T$ token representations and returns a single $d$-dimensional pooled vector, followed by a LayerNorm, concentrating read-out weight on the informative segments of the window. A single linear layer then maps this pooled embedding to the $50$ class logits.

\section{Experiments}
\label{sec:experiments}



\subsection{Dataset}
\label{sec:exp:dataset}

\begin{figure*}[pos=!tbp]
  \centering
  \includegraphics[width=\textwidth]{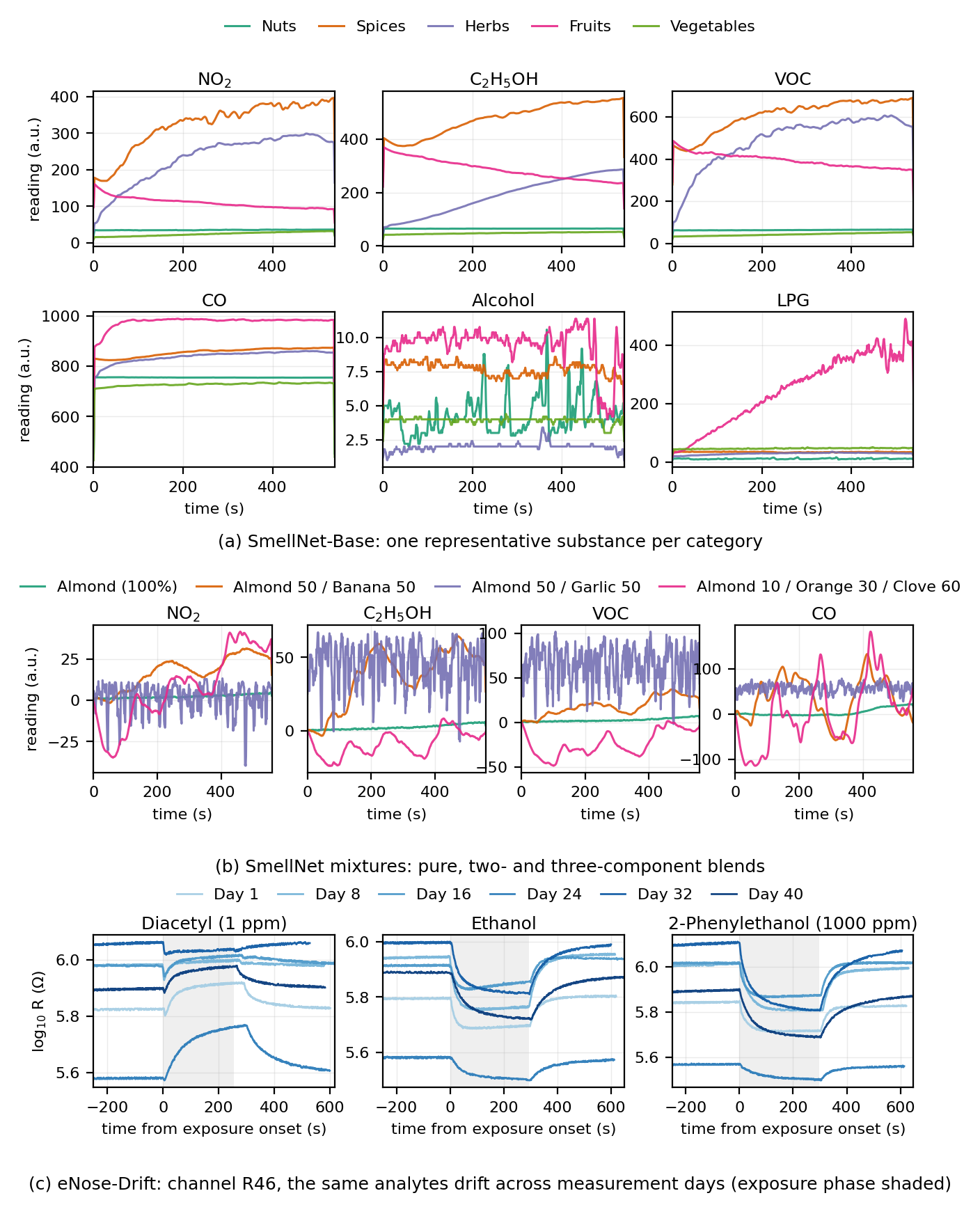}
  \caption{Example sensor time curves from the three tasks. \textbf{(a)}~SmellNet-Base~\citep{feng2026smellnet}: one representative substance per category across the six MOX channels. \textbf{(b)}~SmellNet-Mixtures~\citep{feng2026smellnet}: a pure substance and two- and three-component blends across the four mixture channels. \textbf{(c)}~eNose-Drift~\citep{woerner2025drift}: raw log-resistance of one channel (R46) for the three analytes on six measurement days spanning the year.}
  \label{fig:dataset}
\end{figure*}

Our study uses three tasks (Figure~\ref{fig:dataset}). \textbf{SmellNet-Base}, a 50-way substance-recognition task, serves as the primary benchmark. The model's generalisation is then evaluated on two more challenging tasks, the compositional ratio-prediction task \textbf{SmellNet-Mixtures} and the cross-array recognition task \textbf{eNose-Drift}, which exhibits strong long-term sensor drift.

\textbf{SmellNet-Base}~\citep{feng2026smellnet} is a 50-way substance-recognition task: $50$ base substances spread evenly over five categories (nuts, spices, herbs, fruits, and vegetables, ten each), each sensed by a portable array of low-cost MOX sensors (MQ-3, MQ-5, and the Grove Multichannel Gas Sensor V2) that yields six channels with manufacturer-calibration labels CO, NO$_2$, C$_2$H$_5$OH, VOC, Alcohol, and LPG (not pure-analyte measurements, as each channel is broadly cross-sensitive to many volatile families). Every substance is recorded in six $10$-minute sessions on different days at $1$\,Hz inside a ventilated enclosure, giving one hour per substance and about 50 hours (about 180\,k timesteps) in total. Figure~\ref{fig:dataset}(a) shows example curves of one representative substance per category across the six channels. Following the benchmark protocol, we use six-fold cross-validation at the session level: each fold trains on five recording days and tests on the remaining day, so every day serves once as the held-out test set (about 1\,650 training and 330 test windows per fold at the default $T=300$). Top-1 accuracy on the held-out sessions is the primary Base metric.

\textbf{SmellNet-Mixtures}~\citep{feng2026smellnet} is a ratio-prediction task over twelve base odorants (banana, orange, pear, apple, mango, peach, strawberry, clove, coriander, garlic, almond, and cumin): each recording is a controlled blend of up to three odorants at known proportions encoded in its label (for example a $50/50$ almond and banana blend, or a $10/30/60$ almond, orange and clove blend), captured with the same rig here exposing four channels (NO$_2$, C$_2$H$_5$OH, VOC, CO), and the target is the normalised mixture ratio over the twelve components rather than a class label. Figure~\ref{fig:dataset}(b) shows example curves of a pure substance and of two- and three-component blends across the four channels.

\textbf{eNose-Drift}~\citep{woerner2025drift} is the public \emph{Long-Term Drift Behaviour of an Electronic Nose} benchmark, an independent MOX corpus collected on entirely different hardware. A $62$-channel metal-oxide array repeatedly measures three analytes (diacetyl, ethanol, and 2-phenylethanol) in a baseline-exposure-recovery cycle, yielding about $700$ recordings sampled at $1$\,Hz across $40$ measurement days that span twelve months. Its defining property is strong \emph{long-term drift}: a model trained on early days must stay accurate on much later ones. Figure~\ref{fig:dataset}(c) shows one channel's response to the three analytes on six measurement days across the year. eNose-Drift thus constitutes a rigorous and deployment-relevant evaluation of whether the learnable front-end tuned on SmellNet transfers to a new array, a larger channel count ($62$ vs.\ $6$), a different recognition task, and a pronounced drift regime. We assume only an initial calibration and feed the raw drift-affected log-resistance, adopting a chronological split (earliest days for training, latest for testing) that parallels SmellNet's session-level protocol.

\subsection{Experimental Setup}
\label{sec:exp:design}

We run four experiments to evaluate the method; their results appear in Section~\ref{sec:results}.Unless otherwise stated, raw input means that no external drift correction, smoothing or handcrafted temporal differencing is applied before the network.

\paragraph{Experiment 1: Model Performance} This experiment measures the recognition accuracy of LDAC-Net on the SmellNet-Base task. We report Top-1 and Top-5 accuracy and macro-F1 against four mainstream baselines chosen to span the main temporal inductive biases for multivariate time series: a non-temporal MLP that treats the window as a flat feature vector and ignores order; a 1-D CNN~\citep{lecun1998gradient} that captures local temporal patterns through shared convolutions; a bidirectional LSTM~\citep{hochreiter1997lstm} that models the sequence recurrently; and a Transformer encoder~\citep{vaswani2017attention} that relates timesteps through global attention, the last being the ScentFormer backbone of the SmellNet benchmark~\citep{feng2026smellnet}. Each model is trained on both raw windows and lag-25 FOTD-preprocessed windows. Throughout the paper, \emph{raw input} means the six gas-channel readings exactly as delivered by the benchmark loader: each recording is anchored to its own first sample (the per-session initial calibration, $\mathbf{x}[t]\leftarrow\mathbf{x}[t]-\mathbf{x}[0]$) and sliced into sliding windows, with \emph{no} temporal differencing, filtering, or per-window normalisation applied outside the model; on eNose-Drift the same convention applies to the drift-affected log-resistance $\log_{10}R$, with per-channel standardisation whose statistics are fit on training windows only. We then vary the analysis window length $T\in\{100,200,300,400\}$ at a fixed stride of $50$ to locate the best window length. A longer window plays two roles at once: it regularises training by making neighbouring windows overlap more heavily, and it gives the in-model differencing enough temporal context to be informative. To separate these roles, we run this comparison on two fronts, LDAC-Net and an LDAC-Net w/o LMLD variant (the differencing stage removed), so that comparing the two isolates how much of any gain comes from the learnable differencing rather than from the longer window alone.

\paragraph{Experiment 2: Ablation Study} This experiment attributes the accuracy to individual design choices through a leave-one-out ablation study. Starting from LDAC-Net, we remove WSAN (reverting to plain InstanceNorm), remove the LMLD differencing front as a whole, and remove the LMLD sub-components in turn (the squeeze-excitation gate, the learnable channel mixing, and collapsing the multi-lag bank to a single lag $25$), to quantify each component's contribution.

\paragraph{Experiment 3: Model Comparison} This experiment contrasts LDAC-Net, on raw input, against several of the most popular multivariate time-series models, to show that its gain is not reachable by swapping in an off-the-shelf mechanism. The five comparison methods span the dominant design directions for multi-channel time series, and each is chosen as a strong alternative to one of LDAC-Net's design choices. The Transformer encoder~\citep{vaswani2017attention} models the window with global self-attention and is the ScentFormer backbone of the SmellNet benchmark, so it tests whether plain attention already suffices. The Non-stationary Transformer~\citep{liu2022nonstationary} contributes de-stationary attention, the most direct alternative to our window-conditioned normalisation (WSAN) for handling per-session drift. Autoformer~\citep{wu2021autoformer} brings a trend/residual series decomposition, a different way to separate the slow transient from the response dynamics. The TCN~\citep{bai2018tcn} replaces the attention-convolution stem with a purely convolutional dilated stack, testing the local/dilated-convolution direction. And the Neural-ODE block~\citep{chen2018neuralode} models the window as a continuous-time dynamical system, an alternative to the multi-lag differencing through which LMLD captures the response dynamics. We report Top-1 accuracy for all methods.

\paragraph{Experiment 4: Generalisation} This experiment tests the model's generalisation along two axes: to a different \emph{task} on the same array, and to a different \emph{sensor array} and drift regime altogether. \emph{(a) Compositional generalisation to mixtures} (Section~\ref{sec:res:mixture}) evaluates LDAC-Net on a quantitative, compositional task beyond single-substance recognition. On SmellNet-Mixtures~\citep{feng2026smellnet}, every model emits a $12$-way ratio distribution and is trained with the benchmark's mixture recipe (in-batch synthetic mixing, a KL term, an $\epsilon$-insensitive ratio loss, and a focal presence head). We report three standard ratio metrics: Top-1@0.1 (the fraction of present components whose predicted ratio falls within $0.1$ of the truth), mean absolute error (MAE), and a dynamic Top-$K$ presence score, comparing LDAC-Net against the same four mainstream baselines as in Experiment 1. \emph{(b) Cross-dataset generalisation across sensor arrays} (Section~\ref{sec:res:crossdataset}) evaluates LDAC-Net on the external eNose-Drift benchmark~\citep{woerner2025drift}, which differs from SmellNet in sensor hardware, channel count ($62$ vs.\ $6$), task, and, above all, in exhibiting strong \emph{long-term drift}. We adopt a \emph{chronological} split (train on the earliest days, test on the latest); because the analyte response is a slower transient than SmellNet's, we scale the analysis window to $T{=}400$ and the LMLD bank to $\{25,50,100,150\}$; training runs the same recipe over $100$ epochs. All models receive the \emph{raw} drift-affected log-resistance, with only an initial calibration assumed. We report Acc@1 and macro-F1 (mean$\pm$std over ten seeds for LDAC-Net, the Transformer, and the retuned-FOTD CNN, and over five seeds for the remaining configurations) against the same four mainstream baselines as in Experiment 1, on both raw and dataset-retuned FOTD input.

\subsection{Experimental Configuration}
\label{sec:exp:setup}

\paragraph{Model configuration} LDAC-Net uses model width $d=128$, four attention heads, three AC blocks with FFN width $128$, ConvModule kernel $15$, and dropout $0.1$, per-block stochastic depth $p_{\max}=0.1$, an LMLD mix width $d'=64$, a single-query attention pool, and a WSAN MLP of hidden width $32$. This configuration was chosen by data-scale calibration: every wider or deeper variant we tried overfits. LDAC-Net holds about $7.2\times10^{5}$ trainable parameters in total, of which WSAN adds only about $800$, two orders of magnitude less than ImageNet-class Transformers and matching the order of magnitude of the training corpus. Table~\ref{tab:hparams} lists the full configuration.

\paragraph{Training} The recipe is identical across configurations: AdamW~\citep{loshchilov2019adamw} with learning rate $5\times 10^{-4}$, weight decay $10^{-4}$, batch size $32$, and a cosine schedule~\citep{loshchilov2017sgdr} with $5\,\%$ linear warmup over $100$ epochs. The objective is cross-entropy. Augmentation combines TimeCutout (one contiguous time mask of length up to $T/4$) and Channel Dropout (random masking of one of the six channels); we also use mixup~\citep{zhang2018mixup} with $\alpha=0.2$ (targets interpolated accordingly) and keep an exponential moving average of the weights with decay $0.999$.

\paragraph{Validation protocol} We use K-fold session-level cross-validation, training on five recording days and testing on the remaining day in each fold. Each configuration is run with five random seeds $\{42, 1, 99, 7, 314\}$, and performance is reported as the mean $\pm$ standard deviation.

\begin{table}[pos=t]
  \centering
  \caption{LDAC-Net configuration and training recipe used throughout, unless a row states otherwise. All values are fixed across seeds.}
  \label{tab:hparams}
  \small
  \setlength{\tabcolsep}{4pt}
  \begin{tabular}{ll}
    \toprule
    Setting & Value \\
    \midrule
    \multicolumn{2}{l}{\textit{Data / windowing}} \\
    Channels $C$ / classes        & 6 / 50 \\
    Window length $T$ / stride     & 300 (swept 100 to 400) / 50 \\
    Train / test windows           & about 1\,650 / about 330 \\
    \midrule
    \multicolumn{2}{l}{\textit{Architecture}} \\
    Model width $d$                & 128 \\
    LMLD lag bank $P$ / mix width $d'$ & $\{5,10,25,50\}$ / 64 \\
    Attention heads / blocks       & 4 / 3 \\
    FFN width / ConvModule kernel  & 128 / 15 \\
    Dropout / stochastic depth $p_{\max}$ & 0.1 / 0.1 \\
    Time pooling                   & single-query attention \\
    WSAN MLP hidden width          & 32 \\
    Params (trainable)             & $716{,}746$ \\
    \midrule
    \multicolumn{2}{l}{\textit{Optimisation}} \\
    Optimiser                      & AdamW \\
    Learning rate / weight decay   & $5\times10^{-4}$ / $10^{-4}$ \\
    Batch size / epochs            & 32 / 100 \\
    Schedule                       & cosine, $5\,\%$ warmup \\
    EMA decay                      & 0.999 \\
    \midrule
    \multicolumn{2}{l}{\textit{Regularisation / augmentation}} \\
    mixup $\alpha$                 & 0.2 \\
    TimeCutout / Channel Dropout   & $\le T/4$ / 1 of 6 \\
    \bottomrule
  \end{tabular}
\end{table}

\section{Results}
\label{sec:results}

\subsection{Model Performance}
\label{sec:res:perf}

\subsubsection{Recognition Accuracy on SmellNet-Base}
\label{sec:exp:main}

Table~\ref{tab:main} reports the comparison against the four mainstream architectures: a non-temporal MLP, a 1-D CNN, a bidirectional LSTM, and the Transformer, each trained on raw windows and on lag-25 FOTD-preprocessed windows and scored by Top-1, Top-5, and macro-F1. On raw input every mainstream model is weak. The strongest, Transformer, reaches only $37.29\,\%$ Top-1 ($73.22\,\%$ Top-5, $33.49$ F1), and the non-temporal MLP collapses to $21.09\,\%$ Top-1, because the per-session drift swamps the class signal. FOTD preprocessing recovers much of this gap for the temporal models, lifting the LSTM to $53.78\,\%$ Top-1 ($85.90\,\%$ Top-5, $52.40$ F1), the Transformer to $53.55\,\%$, and the CNN to $49.56\,\%$, but it actively hurts the order-agnostic MLP ($18.92\,\%$). The proposed LDAC-Net, operating end-to-end on raw windows, is the best model on every metric: $\mathbf{68.2\,\%}$ Top-1, $\mathbf{89.4\,\%}$ Top-5, and $\mathbf{65.2}$ macro-F1, exceeding the strongest FOTD-preprocessed baseline by $14.4$ points on Top-1 and $12.8$ on F1 despite using no external preprocessing. This confirms that the drift compensation and dynamics extraction that FOTD performs externally can be folded into the model and surpassed.

\begin{table*}[pos=t]
  \centering
  \caption{Model performance on SmellNet-Base (50-way; Top-1 / Top-5 accuracy and macro-F1, \%, mean$\pm$std). Mainstream baselines (MLP, CNN, LSTM, Transformer) are trained on raw and lag-25 FOTD-preprocessed inputs. LDAC-Net operates end-to-end on raw windows with no external preprocessing. All models use the same analysis window length $T=300$ at stride $50$. Best per metric in bold.}
  \label{tab:main}
  \small
  \setlength{\tabcolsep}{5pt}
  \begin{tabular}{l ccc ccc}
    \toprule
    & \multicolumn{3}{c}{Raw input} & \multicolumn{3}{c}{FOTD input} \\
    \cmidrule(lr){2-4}\cmidrule(lr){5-7}
    Model & Acc@1 & Acc@5 & F1 & Acc@1 & Acc@5 & F1 \\
    \midrule
    MLP          & $21.09 \pm 2.64$ & $56.05 \pm 1.85$ & $16.79 \pm 2.67$ & $18.92 \pm 2.97$ & $49.80 \pm 5.85$ & $16.17 \pm 3.37$ \\
    CNN          & $26.82 \pm 1.44$ & $67.31 \pm 0.65$ & $22.80 \pm 1.99$ & $49.56 \pm 7.80$ & $83.19 \pm 4.03$ & $48.12 \pm 7.20$ \\
    LSTM         & $27.72 \pm 2.40$ & $66.74 \pm 1.17$ & $25.31 \pm 1.73$ & $53.78 \pm 1.81$ & $85.90 \pm 1.14$ & $52.40 \pm 1.99$ \\
    Transformer  & $37.29 \pm 2.34$ & $73.22 \pm 1.73$ & $33.49 \pm 1.99$ & $53.55 \pm 0.83$ & $85.58 \pm 1.86$ & $52.08 \pm 0.77$ \\
    \midrule
    \textbf{LDAC-Net (ours)} & $\mathbf{68.2 \pm 1.6}$ & $\mathbf{89.4 \pm 0.3}$ & $\mathbf{65.2 \pm 4.5}$ & N/A & N/A & N/A \\
    \bottomrule
  \end{tabular}
\end{table*}

\subsubsection{Window Length}
\label{sec:exp:window}

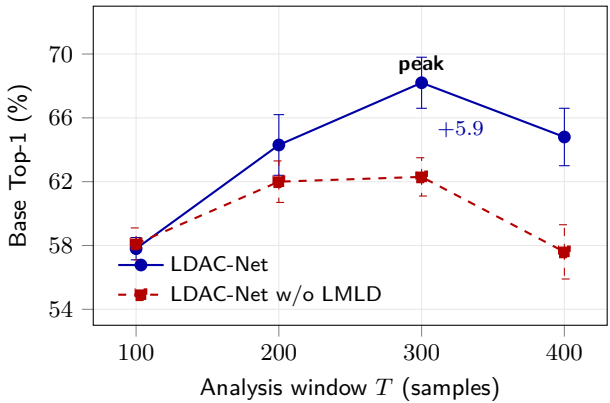
\begin{figure}[pos=!htbp]
  \centering
  \begin{tikzpicture}
  \begin{axis}[
      width=\linewidth, height=5.8cm,
      xlabel={Analysis window $T$ (samples)},
      ylabel={Base Top-1 (\%)},
      xtick={100,200,300,400}, xmin=70, xmax=430,
      ymin=53, ymax=73, ytick={54,58,62,66,70},
      grid=both, grid style={gray!18},
      tick align=outside, tick pos=left,
      legend pos=south west, legend cell align=left,
      legend style={font=\footnotesize, draw=none, fill=none},
      every axis plot/.append style={thick},
      error bars/y dir=both, error bars/y explicit,
    ]
    \addplot[mark=*, color=blue!65!black] coordinates {
      (100,57.8) +- (0,0.7)
      (200,64.3) +- (0,1.9)
      (300,68.2) +- (0,1.6)
      (400,64.8) +- (0,1.8)
    };
    \addlegendentry{LDAC-Net}
    \addplot[mark=square*, color=red!70!black, dashed] coordinates {
      (100,58.1) +- (0,1.0)
      (200,62.0) +- (0,1.3)
      (300,62.3) +- (0,1.2)
      (400,57.6) +- (0,1.7)
    };
    \addlegendentry{LDAC-Net w/o LMLD}
    \node[anchor=south, font=\footnotesize] at (axis cs:300,68.2) {\textbf{peak}};
    \node[anchor=west, font=\footnotesize, color=blue!65!black]
      at (axis cs:305,65.3) {$+5.9$};
  \end{axis}
  \end{tikzpicture}
  \caption{Model performance across analysis window lengths on SmellNet-Base (50-way, Top-1 \%, mean$\pm$std). Solid: LDAC-Net; dashed: LDAC-Net w/o LMLD. LDAC-Net rises by more than ten points from the default $T=100$ to a peak at $T=300$, then falls at $T=400$ as each session yields too few windows. LMLD is neutral at $T=100$, where a lag-$50$ difference is half boundary, and its benefit grows with the window, reaching $+5.9$ points at $T=300$.}
  \label{fig:window}
\end{figure}

In our experiments, we found that the analysis window length greatly affects recognition performance. We vary $T\in\{100,200,300,400\}$ at a fixed stride of $50$ on the 50-way base recognition task, on two fronts: LDAC-Net and the LDAC-Net w/o LMLD variant. Figure~\ref{fig:window} plots the Top-1 mean$\pm$std for both. The classification performance of LDAC-Net improves significantly with the window, rising from $57.8\,\%$ at $T=100$ to a peak of $\mathbf{68.2\,\%}$ at $T=300$ before falling back to $64.8\,\%$ at $T=400$; LDAC-Net w/o LMLD follows the same trend but more weakly, from $58.1\,\%$ at $T=100$ to $62.3\,\%$ at $T=300$ and $57.6\,\%$ at $T=400$. The benefit of a longer window is representational and specific to the differencing front: because a MOX response is a slow chemical transient that takes many seconds to rise and settle, the in-model multi-lag differences need the window to be several times longer than the lag to be informative: at $T=100$ a lag-$50$ difference zeroes half the window, but at $T=300$ only one sixth. The gap between the two curves isolates this effect, i.e.\ the contribution of LMLD: it is negligible at $T=100$ ($-0.3$ points), then widens steadily to $+2.3$ at $T=200$, $+5.9$ at $T=300$, and $+7.2$ at $T=400$. LMLD is not a free addition but one that only a longer window enables. The LDAC-Net curve is unimodal, peaking at $T=300$ and falling at $T=400$ where each session yields too few windows to train on. Based on this, we choose $T=300$.

\subsection{Ablation Study}
\label{sec:res:ablation}

The ablation study is conducted by removing one part of LDAC-Net at a time at $T=300$ and reporting Top-1 (Table~\ref{tab:winner-abl}); every component contributes, starting from the full $68.2\,\%$. WSAN is by far the most important stage: replacing it with plain InstanceNorm drops accuracy to $54.2\,\%$, a loss of $14.0$ points. The size of this drop, from a stage of only about $800$ parameters, matches the mechanism described in Section~\ref{sec:method:wsan}: the two variants differ only in what happens to the window statistics $[\boldsymbol{\mu},\boldsymbol{\sigma}]$, which plain InstanceNorm discards and WSAN routes back through its window-conditioned affine, so the gap measures how much class information the absolute response level carries. The LMLD differencing front accounts for a further $5.8$ points, falling to $62.3\,\%$ when the whole stage is removed, and its internal pieces each leave a clear gap: ablating the squeeze-excitation gate costs $6.2$ points ($61.9\,\%$) and the learnable channel mixing $4.9$ ($63.2\,\%$). The multi-lag bank matters too: collapsing it to a single lag $\{25\}$ costs $4.6$ points ($63.6\,\%$), confirming that differencing at several lags carries information a single lag misses. The dynamic cue LMLD recovers therefore comes from the full combination (differencing across a bank of lags, learnable cross-channel recombination, and gating) layered on top of the window-conditioned normalisation. Every one of these reductions is statistically significant: a two-sided paired $t$-test across the five seeds (which are shared by all configurations) rejects equality with the full model at $p<0.05$ for all five rows, including the single-lag collapse, whose $4.6$-point gap is the smallest in the table.

\begin{table}[pos=t]
  \centering
  \caption{Component ablation of LDAC-Net at $T=300$ (Top-1 \%, mean$\pm$std over five seeds). Each row removes one part of LDAC-Net. All accuracy drops are statistically significant against the full model ($^{\ast}p<0.05$, two-sided paired $t$-test over the five shared seeds).}
  \label{tab:winner-abl}
  \small
  \begin{tabular}{lcc}
    \toprule
    Configuration & Acc.\ (\%) & $\Delta$ \\
    \midrule
    LDAC-Net                              & $\mathbf{68.2 \pm 1.6}$ & (ref.) \\
    \quad $-$ WSAN (plain InstanceNorm)   & $54.2 \pm 0.9$ & $-14.0^{\ast}$ \\
    \quad $-$ LMLD                        & $62.3 \pm 1.2$ & $-5.8^{\ast}$ \\
    \quad $-$ squeeze-excitation gate     & $61.9 \pm 1.3$ & $-6.2^{\ast}$ \\
    \quad $-$ channel mixing              & $63.2 \pm 0.4$ & $-4.9^{\ast}$ \\
    \quad $-$ multi-lag bank (single lag $25$) & $63.6 \pm 2.5$ & $-4.6^{\ast}$ \\
    \bottomrule
  \end{tabular}
\end{table}

\subsection{Model Comparison}
\label{sec:res:comparison}

Table~\ref{tab:mainstream} compares LDAC-Net, on raw input, against the Transformer backbone (on raw and lag-25 FOTD input) and four published time-series methods: the Non-stationary Transformer~\citep{liu2022nonstationary}, Autoformer~\citep{wu2021autoformer}, a temporal convolutional network (TCN)~\citep{bai2018tcn}, and a Neural-ODE~\citep{chen2018neuralode}.

LDAC-Net attains the highest accuracy, outperforming all competing methods by a substantial margin. Among those methods, the de-stationary attention of the Non-stationary Transformer is the strongest at $52.61\,\%$, as expected for the method that most directly stands in for the window-conditioned affine (WSAN), yet it still trails LDAC-Net by more than $15$ points. Autoformer's trend/residual decomposition ($50.28\,\%$) and the Neural-ODE block ($50.47\,\%$) add capacity the small training set cannot absorb, and the dilated TCN stem is the weakest transplant at $43.63\,\%$, about nine points below the best one. The plain Transformer backbone reaches only $37.29\,\%$ on raw input and, even with lag-25 FOTD preprocessing, climbs to just $53.55\,\%$. LDAC-Net at $T=300$ reaches $\mathbf{68.2\,\%}$, more than $14$ points above the strongest competitor (Transformer-FOTD) and over $15$ above the best raw-input alternative. 

\begin{table}[pos=t]
  \centering
  \caption{Comparison of LDAC-Net against the Transformer backbone (on raw and lag-25 FOTD input) and four published time-series methods (the Non-stationary Transformer, Autoformer, TCN, and Neural-ODE) on SmellNet-Base (Top-1 \%, mean$\pm$std).}
  \label{tab:mainstream}
  \small
  \setlength{\tabcolsep}{4pt}
  \begin{tabular}{lc}
    \toprule
    Model & Top-1 Acc.\ (\%) \\
    \midrule
    Transformer (raw)                          & $37.29 \pm 2.34$ \\
    Transformer (FOTD)                         & $53.55 \pm 0.83$ \\
    \midrule
    Non-stationary Transformer~\citep{liu2022nonstationary} & $52.61 \pm 2.50$ \\
    Autoformer~\citep{wu2021autoformer}         & $50.28 \pm 1.22$ \\
    TCN~\citep{bai2018tcn}                       & $43.63 \pm 2.62$ \\
    Neural ODE~\citep{chen2018neuralode}        & $50.47 \pm 1.51$ \\
    \midrule
    \textbf{LDAC-Net}   & $\mathbf{68.2 \pm 1.6}$ \\
    \bottomrule
  \end{tabular}
\end{table}

\subsection{Generalisation}
\label{sec:res:generalisation}

We evaluate generalisation along two axes: to a different task on the same array (mixtures), and to an entirely different sensor array and drift regime (cross-dataset).

\subsubsection{Compositional Generalisation to Mixtures}
\label{sec:res:mixture}

To test whether the representation learned by LDAC-Net generalises beyond the 50-way recognition task, we evaluate it on the SmellNet-Mixtures set, where the goal is to predict the normalised ratio of twelve base odorants in a blend rather than a single class label. Table~\ref{tab:mixture} reports the three ratio metrics, and LDAC-Net is the best model on all three: it raises Top-1@0.1 to $\mathbf{50.5\,\%}$ from the strongest baseline's $45.4\,\%$ (the Transformer), a $5.1$-point gain, while also achieving the lowest MAE ($0.050$) and the highest Top-$K$ presence score ($81.1\,\%$). The learnable multi-lag differencing front and the attention-convolution backbone that drive the recognition result also transfer to a quantitative, compositional task, where they continue to outperform the mainstream architectures.

\begin{table}[pos=t]
  \centering
  \caption{Model performance on the SmellNet-Mixtures task~\citep{feng2026smellnet} (12-component ratio prediction). Top-1@0.1 and Top-$K$ in \%, higher is better; MAE lower is better. Best per metric in bold.}
  \label{tab:mixture}
  \small
  \setlength{\tabcolsep}{3pt}
  \begin{tabular}{lccc}
    \toprule
    Model & Top-1@0.1 $\uparrow$ & MAE $\downarrow$ & Top-$K$ $\uparrow$ \\
    \midrule
    MLP          & $43.0 \pm 1.5$ & $0.056 \pm 0.001$ & $76.2 \pm 1.2$ \\
    CNN          & $43.1 \pm 1.3$ & $0.059 \pm 0.000$ & $75.1 \pm 0.8$ \\
    LSTM         & $41.0 \pm 2.3$ & $0.056 \pm 0.001$ & $80.8 \pm 1.5$ \\
    Transformer  & $45.4 \pm 3.1$ & $0.054 \pm 0.002$ & $78.3 \pm 0.3$ \\
    \midrule
    \textbf{LDAC-Net (ours)} & $\mathbf{50.5 \pm 2.0}$ & $\mathbf{0.050 \pm 0.001}$ & $\mathbf{81.1 \pm 0.2}$ \\
    \bottomrule
  \end{tabular}
\end{table}

\subsubsection{Cross-Dataset Generalisation Across Sensor Arrays}
\label{sec:res:crossdataset}

We evaluate LDAC-Net on the \textbf{eNose-Drift} benchmark~\citep{woerner2025drift}, a $62$-channel metal-oxide array that differs from SmellNet in hardware and channel count. It exhibits strong long-term drift over twelve months (Section~\ref{sec:exp:design}). The four baselines are evaluated on both raw drift-affected log-resistance and dataset-retuned lag-$100$ FOTD input, whereas LDAC-Net operates end-to-end on the raw signal. Table~\ref{tab:crossdataset} reports Acc@1 and macro-F1. On raw input, the baselines attain only $43.4$ to $59.2$ Acc@1 and $38.9$ to $58.1$ macro-F1. FOTD improves the CNN, LSTM, and Transformer in Acc@1, raising them from $59.2$ to $62.6$, $43.4$ to $52.7$, and $56.3$ to $61.2$, respectively, but reduces the order-agnostic MLP from $56.3$ to $53.4$. LDAC-Net achieves the best result in both metrics, reaching $\mathbf{70.6\pm4.8}$ Acc@1 and $\mathbf{69.6\pm3.9}$ macro-F1. It exceeds the strongest raw baseline by $11.4$ Acc@1 points and $11.5$ macro-F1 points, and the strongest dataset-retuned FOTD baseline by $8.0$ and $3.0$ points, respectively. Thus, the model transfers to a new sensor array and drift regime without an externally selected FOTD lag: after scaling the analysis window and lag-bank range to the slower eNose-Drift transient (Section~\ref{sec:exp:design}), its learnable multi-lag front-end recovers the relevant dynamics directly from the raw signal.


\begin{table}[pos=t]
  \centering
  \caption{Cross-dataset generalisation on the eNose-Drift benchmark~\citep{woerner2025drift} ($62$-channel MOX array, chronological drift split, raw drift-affected input). Acc@1 / macro-F1 (\%, mean$\pm$std; ten seeds for LDAC-Net, the Transformer, and the FOTD CNN, five seeds otherwise); FOTD uses the dataset-retuned lag-$100$ difference. Best per column in bold.}
  \label{tab:crossdataset}
  \footnotesize
  \setlength{\tabcolsep}{2pt}
  \begin{tabular}{l cc cc}
    \toprule
    & \multicolumn{2}{c}{Raw} & \multicolumn{2}{c}{FOTD (lag $100$)} \\
    \cmidrule(lr){2-3}\cmidrule(lr){4-5}
    Model & Acc@1 & F1 & Acc@1 & F1 \\
    \midrule
    MLP & $56.3{\pm}4.3$ & $55.6{\pm}4.5$ & $53.4{\pm}1.4$ & $52.4{\pm}1.2$ \\
    CNN & $59.2{\pm}5.3$ & $58.1{\pm}6.7$ & $62.6{\pm}3.1$ & $66.6{\pm}3.2$ \\
    LSTM & $43.4{\pm}3.4$ & $38.9{\pm}3.1$ & $52.7{\pm}1.8$ & $47.8{\pm}2.3$ \\
    Transformer & $56.3{\pm}3.5$ & $55.3{\pm}3.9$ & $61.2{\pm}3.8$ & $58.5{\pm}5.4$ \\
    \midrule
    \textbf{LDAC-Net (ours)} & $\mathbf{70.6{\pm}4.8}$ & $\mathbf{69.6{\pm}3.9}$ & N/A & N/A \\
    \bottomrule
  \end{tabular}
\end{table}
\section{Discussion}
\label{sec:discussion}

\subsection{Window-Conditioned Affine versus Fixed Differencing}

FOTD applies a single fixed map $y[t]=x[t]-x[t-25]$ to every window of every session. Its fixed lag under- or over-compensates sessions whose drift sits at a different time scale, and its differencing discards the absolute response level that is itself discriminative for some classes. WSAN moves the drift-compensation decision inside the model and conditions it on the window's own statistics. A window with a strong DC offset has its mean removed and a learned $\gamma$ rescales the variance, but a window whose absolute level is the discriminative cue can be left near-identity because the MLP $\Phi$ starts at zero and only departs from identity when the data demand it. The ablation in Table~\ref{tab:winner-abl} puts a number on this leaky-normalisation behaviour: the $14.0$-point gap to plain InstanceNorm is the accuracy carried by the absolute-level information that WSAN preserves and InstanceNorm throws away. FOTD's choice of differencing lag is fixed in the same way, and LMLD replaces it with a learnable bank of lags for the same reason. Table~\ref{tab:lagbank} shows why a bank is needed. The single lag that FOTD relies on is the weakest choice ($63.6\,\%$), and adding lags helps only when they span a wide range of time scales: banks confined to the long end ($\{25,50\}$, $\{25,50,100\}$) or to the short end ($\{5,10,15,25\}$), and banks that drop either the shortest or the longest lag ($\{5,10,25\}$, $\{10,25,50\}$), all plateau around $64\,\%$. The bank we adopt, $\{5,10,25,50\}$, is the only one that pairs the fast transient (lag $5$) with the slow rise-and-settle (lag $50$), and it reaches $68.2\,\%$. LMLD learns an operator that adapts to the data's dynamics, weighting and combining differences over a bank of lags spanning the fast transient and the slow rise-and-settle.

\begin{table}[pos=t]
  \centering
  \caption{Effect of the LMLD lag bank on SmellNet-Base (50-way, $T=300$, Top-1 \%, mean$\pm$std). The bank used by LDAC-Net is in bold.}
  \label{tab:lagbank}
  \begin{tabular}{lc}
    \toprule
    Lag bank $P$ & Top-1 Acc.\ (\%) \\
    \midrule
    $\{25\}$               & $63.6 \pm 2.5$ \\
    $\{25, 50\}$           & $64.4 \pm 5.0$ \\
    $\{5, 10, 25\}$        & $64.9 \pm 1.5$ \\
    $\{10, 25, 50\}$       & $64.6 \pm 3.3$ \\
    $\{15, 25, 50\}$       & $64.0 \pm 0.6$ \\
    $\{25, 50, 100\}$      & $64.5 \pm 3.0$ \\
    $\{5, 10, 15, 25\}$    & $63.9 \pm 0.2$ \\
    \midrule
    $\mathbf{\{5, 10, 25, 50\}}$ \textbf{(ours)} & $\mathbf{68.2 \pm 1.6}$ \\
    \bottomrule
  \end{tabular}
\end{table}

\subsection{Window Length as a Key Accuracy Factor}

We find in our experiments that increasing the analysis window length $T$ significantly improves recognition accuracy: lengthening it from the default $T=100$ to $T=300$ raises Top-1 from $57.8\,\%$ to $68.2\,\%$ (Section~\ref{sec:exp:window}, Figure~\ref{fig:window}). This improvement arises for two distinct reasons that happen to align. The first is statistical: at a fixed stride a longer window makes adjacent windows overlap more heavily (each shares most of its span with its neighbours), which acts as a strong augmentation and regulariser even though the window \emph{count} per session falls (from about 2{,}650 windows at $T=100$ to about 1{,}650 at $T=300$). The second is representational: any in-window temporal operation with a lag $p$ (whether fixed FOTD or our LMLD) needs the window to be several times longer than $p$ to be meaningful, otherwise the operation is dominated by the zero-padded boundary. At $T=100$ a lag-$50$ difference is half boundary; at $T=300$ it is one sixth. This is why LMLD is neutral at $T=100$ but its benefit grows with the window: $+2.3$ points at $T=200$, $+5.9$ at $T=300$, and $+7.2$ at $T=400$ in the two-front comparison of Figure~\ref{fig:window}, where the longer window is what gives the differencing front room to work. The curve is unimodal (accuracy falls again at $T=400$ as the window count drops), so there is a genuine trade-off, and $T=300$ sits at the peak.

\subsection{Limitations}

Several limitations remain. First, SmellNet-Base, though the largest controlled MOX benchmark, is still small by deep-learning standards (about $2{,}650$ training windows at $T=100$ and about $1{,}650$ at $T=300$). This scarcity is why every capacity-adding variant we tried overfit and why LDAC-Net is deliberately kept compact. Although the cross-dataset study on eNose-Drift shows the design transfers to a second, larger array, both corpora are modest in size; evaluating LDAC-Net on larger and more varied datasets is still needed to confirm that the gains hold.

Second, the analysis window length $T$, which strongly affects accuracy, is still a global hyperparameter fixed once by an offline grid search ($T=300$), even though the best length almost certainly varies with the substance, the channel, and how fast each response rises and settles. A natural next step is to let the model set it more intelligently, with a learnable, input-adaptive window (for example multi-scale windows), so that the temporal context is selected by the network rather than searched for by hand.

Finally, although the cross-dataset study demonstrates transfer to a second MOX array (eNose-Drift, $62$ channels), both corpora are recorded offline under controlled conditions. Transfer to uncontrolled real-world settings remains untested, and the on-device latency and power of the model, including the small extra cost of the per-window WSAN pass, have yet to be measured in an actual deployment.

\section{Conclusion}
\label{sec:conclusion}

We presented LDAC-Net, a learnable multi-lag differencing attention-convolution network that classifies raw windows from low-cost MOX gas sensors end to end. Its LDFE front end brings drift removal and dynamics extraction into the network. WSAN adjusts the normalisation to each window, while LMLD learns how to combine temporal differences at several lags. The compact AC backbone uses convolution for local response shape and self-attention for longer-range context, with stochastic depth and single-query attention pooling to limit model size. We further investigated the effect of analysis window length. Increasing $T$ from $100$ to $300$ improves Top-1 accuracy by $10.4$ points, from $57.8\,\%$ to $68.2\,\%$. On the 50-way SmellNet-Base task, LDAC-Net achieves $68.2\,\%$ Top-1, $89.4\,\%$ Top-5, and $65.2$ macro-F1. Its Top-1 accuracy is more than $14$ points above the strongest FOTD-preprocessed baseline and over $30$ points above the raw-input Transformer. The ablation study shows that each evaluated component contributes to classification accuracy. Generalisation experiments on SmellNet-Mixtures and eNose-Drift show that LDAC-Net transfers across tasks and sensor arrays. The results indicate that generalisation across recording sessions remains the main limitation, rather than insufficient backbone capacity. They also support learning sensor compensation within the network instead of applying fixed temporal differencing beforehand. We will next test LDAC-Net on larger, real-world multi-session datasets and replace the fixed analysis window with a learnable input-adaptive window.

\ifdefined\ANONYMIZED
\else
\section*{Acknowledgements}
BB/R019983/1, BB/Y513763/1, BB/S020969/1, EP/X013707/1, UKRI3606.

\printcredits
\fi

\bibliographystyle{cas-model2-names}
\bibliography{refs}

\end{document}